\documentclass{article}

\usepackage[final,nonatbib]{midl_2018}

\usepackage[utf8]{inputenc} 
\usepackage[T1]{fontenc}    
\usepackage{hyperref}       
\usepackage{url}            
\usepackage{booktabs}       
\usepackage{amsfonts}       
\usepackage{nicefrac}       
\usepackage{microtype}      
\usepackage{graphicx}
\usepackage{siunitx}
\usepackage{bm}
\title{Adversarial Training for Patient-Independent Feature Learning with IVOCT Data for Plaque Classification}

\author{
  Nils Gessert\thanks{Hamburg University of Technology, Am Schwarzenberg-Campus 3, 21073 Hamburg, Germany.} \\
  \texttt{nils.gessert@tuhh.de} \\
  \And
  Markus Heyder$^*$ \\
  \texttt{markus.heyder@tuhh.de} \\  
  \And
  Sarah Latus$^*$ \\
  \texttt{sarah.latus@tuhh.de} \\
  \And
  David M. Leistner\thanks{Charité – Universitätsmedizin Berlin, Hindenburgdamm 30, 12203 Berlin, Germany } \\
  \texttt{david-manuel.leistner@charite.de} \\
  \And  
  Youssef S. Abdelwahed$^\dagger$ \\   
  \texttt{youssef.abdelwahed@charite.de} \\
  \And
  Matthias Lutz\thanks{Universitätsklinikum Schleswig-Holstein, Arnold-Heller-Straße 3, 24105 Kiel, Germany } \\
  \texttt{matthias.lutz@uksh.de} \\
  \And
  Alexander Schlaefer$^*$ \\
  \texttt{schlaefer@tuhh.de} \\     
}

\begin{document}

\maketitle

\begin{abstract}
  Deep learning methods have shown impressive results for a variety of medical problems over the last few years. However, datasets tend to be small due to time-consuming annotation. As datasets with different patients are often very heterogeneous generalization to new patients can be difficult. This is complicated further if large differences in image acquisition can occur, which is common during intravascular optical coherence tomography for coronary plaque imaging. We address this problem with an adversarial training strategy where we force a part of a deep neural network to learn features that are independent of patient- or acquisition-specific characteristics. We compare our regularization method to typical data augmentation strategies and show that our approach improves performance for a small medical dataset.
  
\end{abstract}

\section{Introduction}

Deep learning methods have been successful for a variety of medical image-based learning problems, such as segmentation or disease classification \cite{Greenspan.2016}. Due to time-consuming annotation, e.g., by a trained expert, medical datasets are often small compared to typical datasets in the natural image domain \cite{Shin.2016}. Often, medical datasets are heterogeneous due to differences between patients which makes generalization to new patients challenging. Variations in data acquisition can complicate this even further. As these variations are independent of the underlying disease, a deep learning model should learn invariance towards these properties, given a large enough dataset. However, for small datasets, a model might quickly overfit to patient- or acquisition-specific features, which requires additional regularization or data augmentation.

This problem is present in particular for intravascular optical coherence tomography (IVOCT) imaging. The technique is used for imaging the inside of coronary arterial walls in order to identify plaque deposits that might lead to atherosclerosis. Deep learning-based plaque classification methods have been proposed, however, their performance is limited \cite{Abdolmanafi.2017}. Thus, data augmentation and transfer learning have been used for IVOCT-based classification \cite{gessert.2018ivoct} which significantly improved performance.

We propose a new regularization method that enforces patient-independent feature learning for small medical datasets. The adversarial training strategy is inspired by domain adversarial training \cite{ganin2016domain}. We adopt the proposed structure and use it for a very different purpose. We show its effectiveness compared to standard data augmentation methods on a small dataset for IVOCT-based plaque classification.

\section{Methods}

\begin{figure}
  \centering
  \includegraphics[width=1\textwidth]{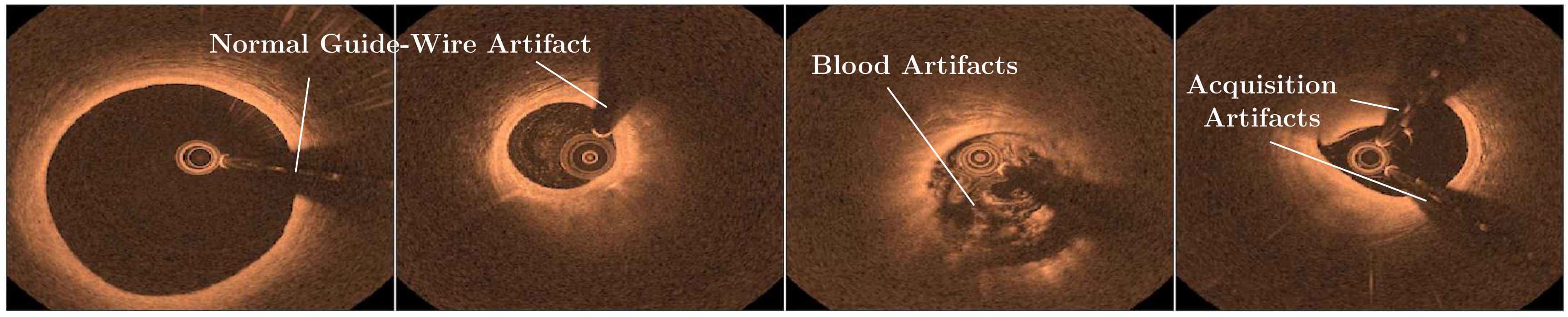}
\caption{Exemplary IVOCT images from different patients in cartesian representation. Note the differences in shape and acquisition artifacts.}
\label{fig:ivoct_example}       
\end{figure}

\subsection{Dataset \& Learning Problem}

We create a new dataset of IVOCT images for plaque classification; as to our knowledge, none are publicly available. The data is acquired in-vivo with a St. Jude Medical Ilumien OPTIS. We use the acquired data in cartesian representation, i.e., the images represent 2D slices of a coronary artery. In Figure~\ref{fig:ivoct_example} we show example images from different patients. Three expert cardiologists with daily routine in IVOCT usage provide the binary ground-truth labels. Each 2D image is assigned a label, "plaque" or "no plaque". In total, the dataset consists of 2600 2D images from 36 patients. For testing, 700 images from 8 patients are split off from that set. In order to show our method's effectiveness with small datasets, we also use a reduced training set size of 20 patients.

\subsection{Adversarial Training Strategy}

\begin{figure}
  \centering
  \includegraphics[width=1\textwidth]{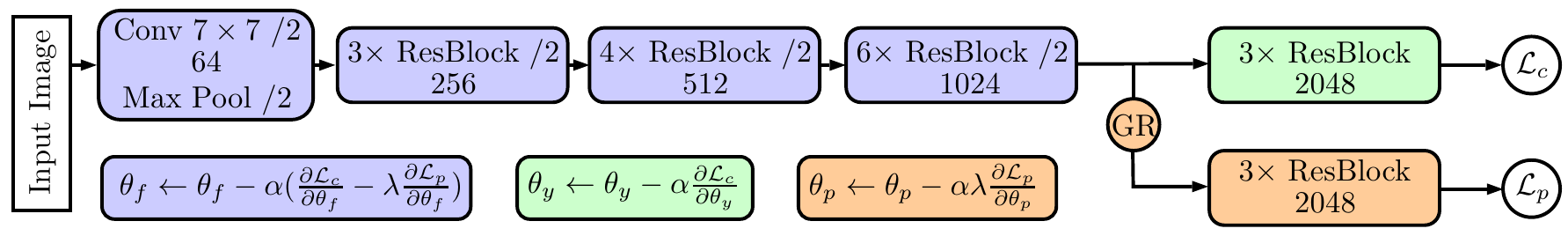}
\caption{Sketch of the used architecture. The number in each block denotes the number of output feature maps. $/2$ denotes a spatial reduction by $2$ within the block. \texttt{GR} denotes the gradient reversal layer. $\theta$ denotes the learnable weights in each part of the model. $\alpha$ denotes the learning rate. Note that the gradient updates are symbolic and the actual updates are performed with the Adam algorithm.}
\label{fig:arch}       
\end{figure}

Our base model is a convolutional neural network (CNN), Resnet50 with post-norm structure \cite{He.2016b}. Similar to \cite{gessert.2018ivoct}, the model is pretrained on the ImageNet dataset. Following \cite{ganin2016domain}, we introduce an additional path, close to the output of the architecture, see Figure~\ref{fig:arch}. The second path is also initialized with the pretrained weights. The main path leads to a binary output that classifies the image with respect to the plaque type present. The second path leads to a classification layer with $N$ outputs, classifying the patient dataset where the image originated from . $N$ is the number of patients in the training set. Using a gradient reversal layer \cite{ganin2016domain}, the loss function
$\mathcal{L}(\theta_f,\theta_y,\theta_p) = \mathcal{L}_c(\theta_f,\theta_y) -\lambda \mathcal{L}_p(\theta_f,\theta_p)$ is minimized which leads to the gradient updates depicted in Figure~\ref{fig:arch}. $\mathcal{L}$ is the cross-entropy loss function and $\lambda$ is a positive scalar which defines the trade-off between the disease and the patient classification loss. We choose $\lambda=0.5$ for our experiments. The two paths compete in an adversarial way where the main (blue) path attempts to learn features which make it impossible to classify from which patient the input image originated. Thus, the main path learns weights $\theta_f$ that maximize the loss $\mathcal{L}_p$ while the second path learns weights $\theta_p$ that minimize the loss $\mathcal{L}_p$, using the features produced by the main path.

We compare this approach to typical data augmentation and regularization. We employ dropout with a keep probability of $p=0.8$. Moreover, we apply random cropping with a crop size of $270\times 270$ from the original image with size $300\times 300$. Additionally, random rotations with $\alpha \in [0\si{\degree},360\si{\degree}]$ and random flipping are applied. 
We implement our models in Tensorflow \cite{Abadi.2016}.

\section{Results}

\begin{table}
	\centering
	\begin{tabular}{l l l l l l l l}
	 & N & AUG & ADV & \textbf{AUG+ADV} & AUG 20 & \textbf{ADV 20} & AUG+ADV 20 \\ \hline \\
	Accuracy & $0.776$ & $0.833$ & $0.873$ & $\bm{0.883}$ &  $0.726$ & $\bm{0.816}$ & $0.805$ \\
	Sensitivity & $0.856$  & $\bm{0.903}$  & $0.857$ & $0.896$  &  $0.740$  & $0.832$ & $\bm{0.859}$ \\
	Specificity & $0.698$ & $0.755$ & $\bm{0.898}$  & $0.872$ &  $0.675$ & $\bm{0.800}$ & $0.752$ \\ \hline \\
	\end{tabular}
	\caption{Comparison of different training scenarios. \texttt{AUG} denotes data augmentation being applied. \texttt{ADV} denotes our adversarial training strategy. \texttt{N} denotes training without either. \texttt{20} denotes training with a reduced training set size with 20 patients.}
	\label{tab:res}
\end{table}

The results for various training scenarios are shown in Table~\ref{tab:res}. Our method shows better results when being applied on top of data augmentation and also without data augmentation. Moreover, the difference in performance becomes even larger when training with a reduced training set.

\section{Discussion \& Conclusion}

We propose a new regularization method for deep learning problems with small datasets. The method uses an adversarial training strategy where the network is forced to learn features that are invariant towards patient-specific characteristics. Our results in Table~\ref{tab:res} show that our method outperforms classic data augmentation techniques for our dataset. When applied on top of data augmentation, accuracy is increased slightly more. However, it is notable that additional data augmentation has little effect when using the adversarial training strategy. This implies that our method already enforces invariance towards properties that are usually covered by data augmentation.
When reducing the training set size, the improvement becomes even more pronounced. This implies that our method prevents overfitting to patient-specific features more effectively than standard augmentation methods.
For further research, our method could be tested with different datasets and more CNN models. 

\bibliographystyle{spmpsci} 
\small      
\bibliography{egbib}

\end{document}